\documentclass[journal]{IEEEtran}
\usepackage{linguex}
\usepackage{natbib}
\bibpunct{(}{)}{;}{a}{,}{;}
\newfont{\eightss}{cmssq8}                      % small sans serif
\usepackage{fancyhdr}
\title{Text Mining for Processing Interview Data in Computational Social Science}
\author{Jussi Karlgren $^{1,*}$ \thanks{$^{*}$ Work done while the first author was a visiting scholar at Stanford, generously hosted by Martin Kay under a VINNMER Marie Curie grant from VINNOVA.}, Renee Li $^{2}$ and Eva M Meyersson Milgrom$^{3}$ \\
$^{1}$  Gavagai, Stockholm\\
$^{2}$  Stanford University}
\date{August 19, 2019}

\begin{document}

\maketitle

\begin{abstract}
We use commercially available text analysis technology to process interview text data from a computational social science study. We find that topical clustering and terminological enrichment provide for convenient exploration and quantification of the responses. This makes it possible to generate and test hypotheses and to compare textual and non-textual variables, and saves analyst effort. We encourage studies in social science to use text analysis, especially for exploratory open-ended studies. We discuss how replicability requirements are met by text analysis technology. We note that the most recent learning models are not  designed with transparency in mind, and that research requires a model to be editable and its decisions to be explainable. The tools available today, such as the one used in the present study, are not built for processing interview texts. While many of the variables under consideration are quantifiable using lexical statistics, we find that some interesting and potentially valuable features are difficult or impossible to automatise reliably at present. We note  that  there  are  some  potentially  interesting  applications for traditional natural language processing mechanisms such as named entity recognition and anaphora resolution in this application area. We conclude with a suggestion for language technologists to investigate the challenge of processing interview data comprehensively, especially the interplay between question and response, and we encourage social science researchers not to hesitate to use text analysis tools, especially for the exploratory phase of processing interview data.
\end{abstract}

\section{Computational social science and text analysis}
Computational social science, as one of its central methods, gathers and processes text data. These data can be obtained through archive studies, media analyses, observational studies, interviews, questionnaires, and many other ways. Computational text analysis has a broad palette of tools to create structured knowledge from human language. The field of language technology, the basis for text analysis tools, has devoted most of its attention to topical analysis of text, to establish what a text or a segment of text is about. This reflects the uses that text analysis has been put to: mostly topical tasks, where relevance and timeliness of the information are the priority. The knowledge gathered from texts in computational social science are, in contrast with the general case, only partly topical in nature, and existing text analysis tools and methods are not always well suited to these types of task. This paper will examine how text analysis can be of use, and point out some cases where current technology could be improved to be more effective. 

\section{Questionnaires and quantifying text data}
There are multiple reasons to gather human data through interviews or questionnaires where respondents can use their own language to respond to queries rather than e.g. multiple choice or graded agreement responses. 

One reason is to make the data gathering situation less formal and more personal and thus encourage the respondent to provide richer data. Gathering data through natural human language allows respondents to express what they \textit{feel}, \textit{perceive}, \textit{believe} and \textit{value}, in a language they are comfortable with which allows the analysis to include the attitude of the respondent to the topic they are responding to. In addition, through an analysis of the language used through the entire interview, general observations about the stance and emotional perspective of the respondent can be made. 

Another reason is to allow the respondent more leeway to formulate their responses without too much imposition from a pre-compiled response structure and thus enable the respondent to provide unexpected data or unexpected connections or dependencies between items under consideration \cite[e.g.]{singer2017somemethodological}. This is especially true if the study concerns (1) a vague or indefinite subject, (2) treats matter for which there is no established vocabulary and phraseology, or (3) which is sensitive in some way \cite{blair1977ask}. An interview situation has the advantage of allowing dynamic follow-up questions which allow specification of aspects that might not otherwise have been detected and will allow the interviewer to probe for bias in the form of tacit assumptions the interviewee might have about concepts that interest the interviewer \cite{lazarsfeld1935artofaskingwhy}. 

For the above reasons, the design of interview studies and analysis of interview data is an area of methodological debate and research within the social sciences \cite{dohrenwend1965some,schuman1979open,wenemark2010respondent}. 

While open answers in surveys and questionnaires provide richer data and reduce the effort and difficulty of formulating the questions in exact form before the fact, they move the effort to after-the-fact-processing of the collected data to get useful results. Open answers are a challenge for analysts: reporting the collected responses together with more quantitative data elicited from respondents is not obvious. Coding procedures---converting open responses into structured form---require time and expertise on the part of the analyst, both of which come at a cost. The effort involved in coding open answers is simultaneously intellectually non-trivial and demanding, but still monotonous: analyst fatigue and frustration risks leading to both between-analyst and within-analyst inconsistencies over time in reporting.\footnote{E.g.  O’Cathain and Thomas \cite{o2004any} and many others; It takes about 1 minute for a human to categorise an abstract, shown by e.g. Macskassy et al. \cite{macskassy1998human} when the categories are already given , If the task is to explore a set of responses and define and revise categories or labels as you go it will involve more effort and require more time per item.}

Freeing human effort from routine tasks to more creative challenges motivates the introduction of computational technology in general, and this field is no exception. Coding textual responses is an excellent example of human effort expended on tasks which are repetitive, time-consuming, and dreary. Processing interview data automatically is an attractive scenario, but is not yet the norm in the social sciences. 

\section{Turning text to quantifiable information}
Language technology has over the last decades developed methods and tools for text analysis, mostly applied to news material, legal documents, technical matters, or other related application domains. The methods have been developed with several characteristics in mind. They are intended to be efficient and effective for practical application to real life problems. They should generalise well from one area to another. In addition, their design is often inspired by human communicative behaviour and as a side constraint it is viewed as desirable in many cases that the models conform to known characteristics of humans process language. 

Text analysis methods rely on observable features found in texts: words, multi-word terms, and constructions. Deciding which features the methods pay attention to is crucial to the quality of the analysis: are words and multi-word terms sufficiently informative to understand the text? Should constructional features, syntactic dependencies, and grammar rules be used? In a statistical model, the chosen features also need to be weighted for importance, e.g. such that an observation of a frequently occurring feature has less weight than an unusual one or that a combination of several features is more interesting than the individual observations. The selected features are then used score texts or text segments quantitatively, to cluster them into coherent subsets, or to classify them into previously established categories. 

All of these three levels of knowledge representation: what features to observe, how they are weighted, and what target categories are of interest, can be learned from text data or explicitly given by a human analyst. That distinction illustrates how text analysis in general spans a spectrum from statistical models to knowledge-based methods. The former, statistical \textit{learning models}, are based on the current generation of machine learning methodologies, learn from examples, and tend to be robust in face of linguistic variation; the latter, by explicitly incorporating previous knowledge from theories of language and lexicography, are able to provide more precise and more sophisticated output. With recent advances in machine learning, these two main approaches are currently converging. 

{\em Text mining} is the general term for the systematic extraction of (somewhat) structured insights from unstructured text material using text analysis methods of various kinds. Text mining in various forms predates the use of machine learning, but most recent approaches are based on learning models. Text mining in general may use any of a range of methodologies starting from simple keyword spotting  to more complex language understanding mechanisms. Text mining also refers to a range of applications, from text clustering and categorisation to more specific analyses, such as extracting items of interest from texts, tailored for some application task. Text mining as an application area traces its roots to the very first steps of language technology, the Linguistic String Project, and \textit{information extraction}, which finds and tabulates pre-identified and carefully formulated structural patterns in text \cite{adams1959office,sager1967syntactic,grishman1996message,grishman1997information, cowie2000information}. In recent years, text mining tools have spread to many practical fields and many overviews of text mining can be found: a technical perspective is given e.g. by Aggarwal and Zhai \cite{aggarwal2012mining}.  

The most popular text analysis method in computational social science currently is the family of methods known as \textit{topic models} which is generally understood to refer to the specific strand of probabilistic models originally defined by Blei, \cite[e.g.]{blei2009topic}. Topic models assume that each text carries some topic or some number of topics selected from a finite set of possible topics. Each topic in play in a text causes some features---terms or constructions, but almost always in practice restricted to single words---to be observable in it. Each observable feature is linked to some topic or some topics with some weight. As an example, texts which can be observed to contain terms such as \textit{helicopter}, \textit{rotor}, \textit{airfield}, and \textit{pilot} can be assumed to be about some topic related to helicopters whereas texts which contain the terms \textit{cow}, \textit{milk}, \textit{cattle}, and \textit{grazing} can be assumed to be about some other topics related to livestock and dairy. The connections between terms and topic and the weights of those connections are learned from observing a large number of texts. In the approach first formulated by Blei, the topics are also learned from texts, without manual intervention, but this does not necessarily need to be always the case: on a more general level, any procedure which relates texts or text segments to a set of topics based on observed linguistic features can be called a topic model, irrespective of algorithmic details. A variant which has been applied to the task of questionnaire processing, \textit{structural topic models}, allow the analyst to insert previous knowledge of covariation between topics into the analysis \cite{roberts2014structural}. The result of a topic model analysis for a text collection is a description of what topics appear in the texts and what features characterise those topics: if analyses on different collections are made, a comparison between them can illustrate e.g. change over time in topic composition.

Topic models have lately been used in digital scholarship for mining historical archives and in media monitoring for mining news feeds and the like. This introduction of new digital methodology for scholarship has not been uncontroversial \cite[e.g.]{fitzpatrick2011humanitiesdonedigitally,moretti2013distant,da2019debacle,underwood2019dearhumanists,da2019computationalcase}. The debate over how to best use new technologies is lively and goes to the roots of what the ultimate research goals of the humanities and the social sciences are. The focus of language technology has been on the analysis of topical and factual content rather than e.g. genre, context, author perspective or stance which has effects on how the functionality is evaluated and developed \cite[e.g.]{karlgren2019lexical}. The concerns of the humanities and the social sciences are methodologically different and they approach knowledge and what insights can be learned from data differently. To some extent, this debate revolves around a technical question: how can technologies that have been developed for some task be transferred to be useful for another? What background assumptions does the technology solution bring to the table and what sort of effects on the results and output of the field does it have?

Computational text analysis has its place in every situation where a collection of text is of a size that would overwhelm a human analyst. The collection does not need to be immense for technology to be helpful. But introducing computational technology where human effort previously has been the norm will have effects on the output, both positive and negative, both predictable and unexpected. For data in social science research, quality criteria go beyond that of productivity gain.

\section{Text mining for the social sciences}

%1. tratt tanken: det allmänna problemet om hur man analyserar survey data  och framförallt semistructured eller unstructured etnografiska explorativa data. Det specifika problemet att använda algoritmer som ex topologiska modeller, och det superfokuserade problemet, hur långt kommer man med de existerande modellerna, vilket ju är vår studie antar jag, vilka är de existerande topologiska modellerna tillkortakommande och vad behöver vi utveckla?

Text analysis in questionnaire and interview studies involves establishing whether the text verifies or refutes hypotheses of interest, or exploring the data to establish hypotheses for continued study and testing. In practice, this means finding if some concepts or entities of interest are mentioned in the text data and to establish how often, in what contexts, and in what way they are mentioned. Processing questionnaire and interview texts will entail establishing what general topic is being discussed, but also tracking mentions of e.g. persons, organisations, relations between them, and attitudes and feelings with respect to them. Potentially, the analysis is able to make use of background variables such as frame of mind, personality type, and outlook of the respondent. 

One of the most valuable features of human language is that it allows its users to vary their way of expressing themselves depending on the social context, the attitude and background of the speaker, and many other factors. This is a challenge for analysis methods that rely on observations of term occurrences. Most concepts or notions of interest can be referred to with a variety of terms, and most terms may, depending on context and other variables, refer to a multitude of concepts. The ambiguity, vagueness, and fluidity over time and across situations of human linguistic expression is often described as a problem. This perspective does not do justice to the nature of human communication. The adaptability of human language is useful: it allows new terms to be coined, established terms to be recruited into service to fit the momentary needs of some discourse, and various discourses to be associated or contrasted through choices of terms and constructions. The challenge for the analyst of our specific use case is in fact exactly the reason why open answers are useful: if the choice of terms and constructions were entirely predictable, the information captured through open answers would be so much less rich and valuable.

For research tasks, \textit{replicability} of an analysis is a major quality criterion. One must be able to revisit the data and reproduce the analysis results using the methodologies originally used for analysis. Furthermore, researchers also wish to generalise from a given study to new populations. This is as true for text data as it is for other types of data. Text classification and processing is today practically always done by the analyst or researcher in person, or by hired human coders. For these scholars to trust computational technology the tools must provide replicable, consistent, and explainable results.  

Given the dynamic and situational characteristics of human language, matching interesting concepts and relevant research variables to observable features in interview responses is a non-trivial challenge. Doing this reliably and consistently encompasses every challenge in language understanding. It involves terminological, situational, social, dialectal, and individual variation. The degrees of freedom are obviously larger for explorative studies where hypotheses are unformed and the concepts are not defined. Classifying texts and identifying concepts in responses reliably requires that the classifier has seen large amounts of text in general and has an understanding of the topic being discussed. Human analysts have this competence as a matter of course, but text analysis tools need to acquire the competence somehow. 

Recent approaches in text analysis rely on combinations of learning models. They are versatile and trainable for application in fields where computational models previously have not been of use. Learning models in general are built by providing them with \textit{training data} which are used to set their internal parameters, weights, or loadings in a way which allows them to repeat their performance on previously unseen data. There are numerous learning schemes which rely on various levels of manual intervention and human effort to guide them. The practical objective of introducing learning models is to reduce the amount of human effort in configuring a tool to handle the specifics of human language in a new domain. 

Firstly, in recent years, \textit{end-to-end training} implemented using \textit{neural models} has become the dominant framework for learning models. Neural models are not designed to be configured by human effort, but are instead implemented to learn from examples with as little intervention as possible. They are inspired by biological neural architectures and are implemented using a large number of internal statistical parameters which are purposely hidden from the human operator, instead to be set and modified through exposure to judiciously chosen training examples. A neural model is trained by being shown a number of training texts together with manually assigned desired category labels. It will then use that information to select combinations of observable features in the training texts, for use for future categorisation of previously unseen texts. This is a convenient way to build a statistical prediction model with little technical effort, and this combination of approaches has proven useful in many text analysis tasks with impressive performance in recent years. The end-to-end training framework corresponds to giving human analysts example results to use as a model for the analysis of new data rather than instructing them by giving them rules for how to proceed. 

Secondly, the background understanding of language is learned through the \textit{transfer learning} framework, by processing large amounts of previous text and applying that knowledge to numerous later analysis tasks. These previous texts do not need to be from the same domain, since the objective is to learn general statistical patterns: which terms and constructions are frequent and which are unusual, e.g. By providing a base from which to inspect new data, the transfer learning framework enables learning models to more rapidly establish what is interesting in the data set under analysis. This corresponds to requiring human analysts to be well read and erudite. 

Tireless processing at high speed is an obvious benefit of using computational approaches, but answering to the requirement for replicability needs attention before computational models can be reliably applied to texts in social science. Statistical models generally and neural models specifically are designed to hide their parameters from the human operator. Even if internal parameters were to be inspected, they would be meaningless, taken individually, since neural models encode knowledge as patterns over a multitude of statistical scores rather than isolated in an individual variable. This is one of the main advantages of neural models, since it allows the model to take many observable features into account simultaneously, even such features that are weak taken in isolation. Such cooccurrence data are impossible to enumerate manually, but can be learned if the training data set is large enough. This has the consequence that neural models rely heavily on appropriately chosen training data. This is where the transfer learning framework helps by leveraging previously collected data. Those data contribute to the analysis in various ways, but what their effect is can only be established if the model comes with a declaration of what data they were trained on. These qualities of neural models and the transfer learning framework contribute to making learning text analysis models less \textit{transparent} than a model which would rely on e.g. a set of written explicit categorisation rules. This is a known challenge in many applications, and is partially addressed in recent research to make artificial intelligence and learning models explainable\footnote{E.g. in the Explainable Artificial Intelligence (XAI) program launched by DARPA in 2016.}. As an additional facet of transparency, for application to research tasks, it is desirable that the model be {\em editable} to allow the researcher to manually improve its precision and recall over the data set under consideration for cases where the automatic mechanisms may have not found some correspondence of interest or where they may have overtrained on some singularity in the data set. This is in contrast with how end-to-end models typically are understood, but if the aim of a study is not to automatise concept learning but to produce an outcome to be trusted, professional human intervention is an important part of the workflow. The structural topic models mentioned above are one way of moving in this direction, since they allow for the explicit introduction of e.g. expected covariation between topics \cite{roberts2014structural}. As a general requirement for using learning models in a research task, models should allow the analyst to inspect the classification or scoring it provides and allow the analyst to adjust internal parameters to fit hypotheses or established knowledge. This is not always the case for current learning models.

\section{Case study}
The case study which motivated us to address these more general methodological questions is an ethnographic exploratory study---"Intimacy after 60. Transition into retirement"---on women's experiences and current feelings about relationships and intimacy during their transition from working life to retirement. The study investigates how mindset and attitude towards relationships with partner, family, friends, and colleagues, with respect to compromises, principles, and conflict resolution relates to how the subjects value three aspects of intimacy: physical, emotional, and intellectual. The case study is explorative and is intended to generate patterns which will be used to formulate hypotheses for future testing in representative longitudinal panel studies and the exploratory design was formed with that in mind. 

Recordings of these interviews have been manually transcribed. Each interview in the study consists of a number of \textit{turns}: the interviewer asks a question or prompts the respondent in one turn and the respondent then reacts to the prompt in another turn, ranging in length from terse answers (\ref{terse} to paragraph-length multi-sentence responses (\ref{verbose}, \ref{married}, \ref{careof}). We use turns as the basic unit of analysis. 

\ex.\label{examples}
\a. "I forgot to mention them. They're another two that are best friends, yes. {\sc Name1}, and {\sc Name2}, and I are very close. "
\b. "I think he was not a well person, because he always managed to arrange it so that I would find out. Maybe, you know, a piece... "
\c."Yes I was"\label{terse}
\d.There is something that I talk about only with one friend, although a couple of friends know about it. Yes, but it's funny you should say that--I was recently spending a weekend with friends, {\sc Name1} and {\sc Name2}. And {\sc Name1} and I are very close. {\sc Name2} and I are close, but he and I are closer. And we were walking on the beach--so tranquilizing, the water--and he said to me, he said, "You know, it's the stories that they don't tell about ourselves that are the ones that really define us." \label{verbose}
\e. "..."Oh, well, this is forever, and it'll just be the two of us." I think we got married so young because we so badly wanted to be together, but in those days it wasn't really very nice to be sexual unless you were married. So... " \label{married}
\f. Not necessarily being taken care of--although my friends always want to be very careful with me and kind--but I think what I want from the relationship is to not be alone, whether it's intellectual, you know, and we spend an hour bashing the president as we did the other night, or, you know, a common interest. I have friends that I go birding with, you know? I think it's just important not to feel all by myself, not to feel too abandoned. \label{careof}

Many variables are obviously given directly by study design and respondent selection, through various metadata (e.g. demographic and socioeconomic variables). As an example, the data collection for the study has been performed in a series of interviews in several cultural areas: North America (NA), Northern Europe (NE), Asia (A), Latin America, Eastern Europe, and the Middle East. Countries characterised by recent political upheavals, lack of social peace, or weak rule of law have been grouped under (W). The data set currently under consideration consists of some 54 interviews. Overview descriptive statistics are given in Table~\ref{culturestats}. While the scale of this study is less than typically is thought to be of interest for automatic processing, even this collection of turns is time-consuming and daunting to process consistently for a human analyst. 

\begin{table*}[htpb]
\caption{Descriptive statistics for the collected data.}\label{culturestats}
\centering
\begin{tabular}{r|rrrrr}
\hline
Cultural area		&	Interviews	&	 Words	&	 Turns 	&	Turn length 	&	Interview length	\\
 & & & & {\tiny (average number of words)} & {\tiny (average number of words)} \\
\hline
NA	&	20	&	88~208	&	3~631	&	24.3	&	181.6	\\
A	&	5	&	21~109	&	1~006	&	21.0	&	201.2	\\
W	&	14	&	2~746	&	1~282	&	18.6	&	91.6 \\
NE	&	15	&	61~456	&	2~345	&	26.2	&	156.3	\\
\hline
\end{tabular}
\end{table*}

Text analysis is used in this study to identify a number of observable textual features which can be quantified into relevant variables for the study. The language used by the respondents is not a one on one mapping onto variables. Observable features such as term counts need to be combined and sometimes weighted to be aggregated into useful variables. Some concepts are mentioned in several turns; some attitudes only emerge across the entire interview; some concepts are only implicitly present and can be inferred by a human reader with some accuracy but are yet beyond the capability of automated methods. 

The tool used in this present case study---Gavagai Explorer---is built primarily for interactive analysis of e.g. customer feedback, consumer reviews, or market surveys but has also been used in previous academic research, \cite[e.g.]{georg2019virtual}. Its functionality is based on term statistics: it uses words and multi-word terms as features to cluster texts. Gavagai Explorer is built to be transparent: the models used in the analysis are represented as editable lists of words and terms, which allows the analyst to modify the clustering criteria and recluster interactively. Gavagai Explorer uses extensive background knowledge of general language usage based on large amounts of continuously ingested streaming text from published materials, both editorial and social media, and is thus an example of the transfer learning approach discussed above \cite{sahlgren2016gavagai}. Gavagai Explorer splits each text item---in this case, an interview turn---into sentences and clusters those sentences into topically coherent sets by terms that occur in them. Each sentence can only be in one cluster, but each turn, since it may contain several sentences and treat several topics, can be in several clusters. The clusters and the terms that define them can be inspected and edited by the analyst by adding or deleting terms to enrich and refine them. Clusters of no or little interest can be merged with others or discarded entirely. Using the background knowledge of language usage, the analyst is given suggestions of suitable synonym or related terms. In addition, each turn and each cluster is scored by a number of sentiments which can be defined as sets of terms by the analyst \cite{espinoza2018analysis}.

\subsection*{Codable variables} 
Some variables of the study are given as direct answers to questions posed by the interviewer.  With direct answers of the type shown in Examples \ref{codables}-\ref{codablescale}, there is a clear and codable datapoint to be found in the response. In many cases, the topic is only mentioned in the question such as in Example~\ref{codableyesno}: responses can be short and abbreviated and refer back to the question through implicit continuation along a topic introduced by the question. A question does not reflect the language of the respondent, but text from a question-response pair---and in some cases, from an even longer stretch of discourse---may necessary for the analysis of what the respondent means. 

These responses are straightforward for human coders to extract from the material, but are still fairly tricky for automatic analysis. In the present study, such variables have been analysed by hand. Automatically processing these kinds of variables would be possible with high precision for many of the cases, but extracting them automatically will entail loss of coverage, using a combination of information extraction and recent advances in sequence tagging with machine learning models. Today, the general case is not yet resolved, and the special cases that can be solved will need large amounts of training data to attain any level of reasonable coverage. Processing question-response pairs in a longer discourse is today addressed in the context of conversational agents in e.g. customer service contexts. That methodology shows promise of being applicable for the purpose of processing interviews. 

\ex.\label{codables}
\a. INTERVIEWER: Did you work full-time?
\b. RESPONDENT: I always worked full-time except for when I was in school, yeah.

\ex.\label{codableyesno}
\a. INTERVIEWER: Do you belong to any religious community?
\b. RESPONDENT: No.

\ex.\label{codablescale}
\a. INTERVIEWER: On a scale from 1 to 10, how much did your parents encourage you to get an education?
\b. RESPONDENT: Not so very much.  I'm going to say 4.

\subsection*{Reference to concept}
The variables of greatest interest for this present discussion are those expressed by the respondents in free form in unconstrained discourse. Tabulating these can be done through analysis of respondent turns in the text, and the freedom they afford the respondent are the reason to move to open responses in the first place. Much of this is fairly simple to detect. If someone mentions \textit{partner}, we know they are talking about someone's partner, and given the interview situation this is almost certainly the respondent's partner. By simple tallying of how often some concept is mentioned we get a quantitative score of its importance to the subject in question, and this score can be compared to other respondents' scores. 

This fits well with the goals of general text analysis and especially that of topic modelling discussed earlier. Features, especially terms, that are observed to occur frequently are likely to be important in a text, but only if they occur more frequently than expected; especially interesting are features that occur burstily, with local peaks in distribution, indicating that some matter of interest is under treatment \cite{katz1996distribution,madsen2005modeling}: how to compute and assess the importance of term occurrence is central to the theory of information retrieval and related technologies \cite[e.g.]{luhn1957statistical}. This reflects one of the more important aspects of language use: that of \textit{referentiality}, where language is used to refer to entities, concepts, or notions of interest to discourse participants. Referring to the dynamic and flexible qualities of human language discussed above, reference to concepts can be done in many ways, some not entirely predictable before the fact. To ensure \textit{recall}, or coverage, of the analysis, semantically related terms must be included: for the \textit{helicopter} example given earlier, we might want both synonyms or near synonyms (\textit{autogiro, chopper, whirlybird}) and other related terms (\textit{airfoil, camber, translational lift}) to make sure we identify every time the respondent addresses a concept; for the \textit{partner} example in this case study, we would at a minimum wish to include e.g. their \textit{husband}, their \textit{boyfriend}, their \textit{ex} etc, perhaps extending to \textit{married}, \textit{engagement}, and so forth. This is where transfer learning is of great help: manual enumeration of related terms by the analyst would be difficult to do exhaustively and reliably. By using occurrence statistics from general language use it is relatively easy to identify and suggest semantically related and associated terms which yields a model with greater consistency and coverage alike.  

The concept clusters for this present case study include aspects of closeness such as \textit{intellectual closeness}, \textit{emotional closeness}, \textit{physical closeness}, \textit{sexual closeness}; as well as various interpersonal relations of the respondent such as \textit{partner}, \textit{friends}, \textit{colleagues}, and \textit{family}. These are examined and iteratively refined by editing the terms that define them, resulting in sets of terms of up to a hundred in size. Samples of two automatically grouped concept clusters are shown in Examples~\ref{travelcluster} (clustered around various leisure activities) and \ref{palcluster} (clustered around the concept of friendship).

\ex.\label{travelcluster}
\a. I love theater and movies.
\b. I travel with my daughter, my cousin, and the last trip I did I did alone.
\c. Movies I also like and I do knit.
\d. Netflix is wonderful, we have a glass of wine and peanuts and a movie.
\e. He did lots of traveling.
\f. I do yoga two times a week and I dance three times a week and I do gymnastics.
\f. We do concerts and movies together.

\ex.\label{palcluster}
\a. With my friends I had more freedom.
\b. Lately, I let go of friends that don't work out anymore.
\c. Different groups where I have my best friends.
\d. I talk about my issues with my best friends.
\e. Now he has a girlfriend from {\sc Place}
\f. I am intellectually close to my friends and also emotionally.
\f. I really enjoy talking to my friends about their experience of politics, my friend used to live in {\sc Place}, he lived there for years during Pinochet.

Using counts of how often concepts are mentioned, we are able to assess the relative importance of the various types of closeness related to the differential between the various cultural areas, and an overview is given in Table~\ref{cultureintimacy}. The table shows what proportion of the turns bring up the aspect of intimacy in question. We see that respondents from Northern Europe do not address \textit{physical closeness} or \textit{intellectual closeness} as often as other respondents do, and that respondents from North America discuss \textit{sexual closeness} more often than others. 

\begin{table*}[htpb]
\caption{Difference in emphasis on the various aspects of intimacy across cultural areas (percentage of all turns that bring up the facet of intimacy in question)}\label{cultureintimacy}
\centering
\begin{tabular}{r|rrrr}
\hline
\textbf{Cultural area}	& \textbf{Physical}	& \textbf{Emotional}	& \textbf{Intellectual} & \textbf{Sexual} \\
\hline
all     &	6.34\%	&	6.64\%	&	4.85\%	&	2.61\%	\\
\hline
NA	    &	8.32\%	&	5.98\%	&	5.34\%	&	3.30\%	\\
A	&	6.86\%	&	6.46\%	&	5.57\%	&	1.49\%	\\
W	&	9.44\%	&	8.11\%	&	6.90\%	&	2.50\%	\\
NE	&	1.36\%	&	6.95\%	&	2.69\%	&	2.09\%	\\
\hline
\end{tabular}
\end{table*}

There is an additional referential pattern that has not been captured by the current analysis. Coherence in a conversation may be realised using e.g. pronominal reference to a previously named person. Example~\ref{pronoun} shows how the response never mentions "husband". Interpreting the turn requires the analyst to refer to the question to resolve who is posing the budgetary requests under discussion. Similarly, reference by person names is frequent in this material, as in Example~\ref{ner}, where the respondent uses the first name of a previous husband to refer to him.

\ex.\label{pronoun}
\a. INTERVIEWER: Was it the same with your first husband?  Was he supportive of you working full-time when you had kids?
\b. RESPONDENT: Well, when I had---when we moved to {\sc Place} in 1989, we didn't have a lot of money, and I would say that he wanted me to do more work, because I wasn't bringing in money.

\ex.\label{ner}
\a. INTERVIEWER: And when you were 23 you lived for how long with ...
\b. RESPONDENT: With {\sc Name}? Er, altogether maybe two or three years

Pronoun resolution---figuring out who "he" or "him" refers to locally in discourse such as in Example~\ref{pronoun}, is a known and on a theoretical level solved task. In this case, we cannot trust such algorithms to resolve the "he" to the right referent (are we now discussing the partner, a son-in-law, a co-worker?) Similarly, identifying person names is in theory a similarly simple challenge: named entity resolution in general case is a solved task in language technology. However, in cases such as in Example~\ref{ner}, resolving who {\sc Name} of the various candidate persons mentioned during the interview is involves some knowledge of the limits of the discourse at hand. Recently introduced sequence tagging models promise even better coverage than previous knowledge-based models and addressing this challenge promises to be a very fruitful avenue of near future research.

\subsubsection*{Attitude, mood, and characteristics of the respondent}
Other aspects of language use are not as obviously calculable by examining simple term counts, such as those that encode \textit{relations} between notions that are referred to, those that organize the \textit{structure} of the discourse, and those that indicate speaker or author \textit{attitude}, \textit{stance}, or \textit{mood}. These latter aspects relate directly to variables of interest for computational social sciences: in the case study at hand, e.g.  \textit{tentativeness} vs \textit{executive capacity}; \textit{pragmatic} vs \textit{principled} mindset; \textit{empathetic} vs \textit{solipsist} outlook and other similar personally or culturally bound concepts. There is no obvious and crisp definitional demarcation between features that refer to concepts and features that communicate e.g. attitude, but there are some observable differences in distribution. Features of the latter kind are sprinkled throughout the text data and cannot be pinpointed to any single utterance. In addition, the distribution of attitude features dispersed across texts than features that refer to some concept: \textit{skepticism} is more frequent in language in general and expressed here and there than terms that refer to e.g. helicopters. This poses a challenge for computational approaches such as topic modelling which have been optimised to establish what concepts a text refers to. Even so, human readers are able with some precision to distinguish many such linguistic factors from reading text, which should be understood to mean that text analysis should be able to establish observable features for them. Work on \textit{sentiment analysis} which focusses on detecting positive vs negative sentiment in text is one example where such variables are detectable \cite[e.g.]{liu2012sentiment}. The palette of human emotion is more complex than so: a text can be written with skepticism, revulsion, anger, or frustration, and what the composition and parameters of variation of human emotion are is under continuous discussion \cite{darwin1872, james1884, meh1974,  morganheise1988, ekman1992, kuppensEA2004, handbook2007, diff2009} and attendant development of methods to detect sentiment in text. In a study such as the present example, some expressions of sentiment are pertinent to the research hypotheses. Example~\ref{negativity} gives some samples of items with negative sentiment from the interviews. In this study, we defined expressions of feelings and sentiments to suit the hypotheses of the study such as "Feeling rejected" or "Taking initiative" to be valuable addition to standard positivity and negativity. These expressions are defined as lists of terms, which ensures transparency and editability of the model. 

\begin{table*}[htpb]
\caption{Cultural areas and attitude in text (scores computed based on weighted occurrence of polar terms, attitudinal terms, and hedge terms respectively) }\label{culturalpersonality}
\centering
\begin{tabular}{r|rrr}
\hline
\textbf{Cultural area}	&	\textbf{Polarity}	&	\textbf{Intensity}	&	\textbf{Skepticism} \\
\hline
NA	&	26.8	&	193.0	&	11.1	\\
A	&	21.6	&	163.3	&	11.0	\\
W	&	11.8	&	256.9	&	2.64	\\
NE	&	10.1	&	66.3	&	22.8	\\
\hline
\end{tabular}
\end{table*}

\ex.\label{negativity}
\a. I'm so very sorry. I didn't expect to be talking about all this horrible stuff. Thank you.  
\b. I'm sad hearing that. The first one I think was okay, but the second two... Because then you're using sex...
\c. Of course they treated her terrible, right?
\d. I believe that he married her to have access to the children, because he was a monster.
\e. So I became miserable there.
\f. But he verbally abused me horribly.
\f. It was terrible, it was terrible, and nobody knew.  

The model allows us to define variables over all the scores from a respondent, to assess their individual qualities. Some of the hypotheses of the current study call for e.g. distinguishing pragmatic respondents who put effort into compromise and conflict resolution from respondents who hold their social circles to set standards and whose social actions are bound by explicitly formulated principles; respondents with a positive outlook from those who are more gloomy; respondents who express emotions with intensity from those who are more reticent; and other similar personally or culturally bound concepts. In contrast with sentiment, which is here understood as an attitude shown vis-à-vis some mentioned topic, mindset markers are sprinkled throughout responses, cannot be found in any single turn, and need to be aggregated over an entire session. If a respondent repeatedly uses terms to indicate bitterness, uncertainty, or exuberance, these can be aggregated.  Recent interest in authorship profiling has provided some tentative feature sets to explore personality type and various demographic variables from observable features \cite{rangel2013overview}: in this study we have used the attitude scores to compute the difference between number of observed expressions for various positive sentiments and for various negative sentiments as a measure of \textit{polarity} of emotion for individuals and a sum of squares for both as a measure of \textit{intensity} of emotion for individuals. 

The variation over aspects of emphasis shown in Table~\ref{cultureintimacy}, and using attitudinal variables we are able to further analyse it to examine differences in attitude. In Table~\ref{culturalpersonality} we find that hedged or cautious expressions with overt markers of skepticism are more prevalent and intensity of expression is less pronounced in responses from Northern European respondents than in others and that cultural areas where political upheaval has been present show the opposite pattern.

Across the entire data set, for all individuals, we find e.g. that placing high importance on physical closeness is well correlated with placing high importance on intellectual closeness, but that importance of emotional closeness correlates less with the other facets of closeness. We also find that those who express themselves with more positive than negative terms more often bring up the concept "Getting along" than others.

\section{Conclusions, Lessons learnt, and Paths Forward}
We have used text analysis technology to process interview text
data from a computational social science study. 
The tool we used to process the text data is developed for the analysis of
customer feedback and related data: a task similar enough
to processing data for the social sciences.

We found that topical clustering and terminological enrichment provide
for convenient exploration and quantification of the responses:
tabulating observed concepts allows e.g. for correlation analysis and
inference of relations between demographic data and attitude towards
concepts of interest for the study. This makes it possible to simultaneously establish both textual and non-textual variables and to generate hypotheses
for further testing. Introducing computational tools saves analyst effort even with a text collection on this small scale.

The tools available today, such as the one used in the present study, are not tailored to the task of processing interview texts. We find that some interesting and potentially valuable features are difficult or impossible to automatise reliably at present. We especially note as potentially quite useful avenues of further investigation that:
\begin{enumerate}
    \item Reference to concepts of interest is relatively straightforward to establish whereas attitudinal content in many cases poses greater challenges. There is a large and growing body of work on attitude and sentiment analysis in text analysis, but in most cases, the palette of emotions or attitudes will need to be tailored to the needs of the study at hand, which will require more work on the part of the analyst and requires the tool to accommodate such editorial work. 
    \item Variables that are distributed over the entire text of an interview are more challenging to extract: in this case study, e.g., establishing the mindset of the respondent with respect to their social circles, whether they are principled and rule-bound or whether they tend towards compromise and pragmatism. Automating such analyses using recent machine learning models is again quite possible, but needs new learning components tailored for this purpose. 
    \item The interplay between questions and responses poses specific requirements for text analysis which are possible to address using today's technology if correspondences of interest are used to train a model. This has to our knowledge not yet been attempted.  
    \item Traditional natural language processing mechanisms developed for general application in text analysis such as resolving referents for proper names and pronouns are excellent candidates to improve the analysis, but need to be tailored to the specifics of question-response interplay. 
\end{enumerate}

Using computational text analysis tools for research in social science has great potential to allow for more exploratory open-ended studies with less effort, increase coding consistency, and reduce turnaround time for the analysis of collected data by using tools developed for market purposes.

We note that any analysis model used for research purposes where previously human effort has been the norm, must provide \textit{replicable} results. To be replicable, the deliberations of a tool must be \textit{transparent}, to ensure quality, the knowledge it uses must be {\em editable}, and to be acceptable for research, classification decisions and scores must be {\em explainable}. These requirements are necessary both to afford the researcher trust in the analysis results and for review, replication, and reuse by others. Many recent methods and tools are not built with these criteria in mind.

We conclude by urging researchers in language
technology to investigate the specific challenges of processing interview
data, especially referential issues such as the interplay between questions and responses, and we encourage social science researchers not
to hesitate to use text analysis tools, especially for the exploratory
phase of processing interview data.
\newpage

\bibliographystyle{unsrtnat}

\bibliography{textanalysiskarlgrenlimeyerssonmilgrom}

\begin{thebibliography}{38}
\providecommand{\natexlab}[1]{#1}
\providecommand{\url}[1]{\texttt{#1}}
\expandafter\ifx\csname urlstyle\endcsname\relax
  \providecommand{\doi}[1]{doi: #1}\else
  \providecommand{\doi}{doi: \begingroup \urlstyle{rm}\Url}\fi

\bibitem[Singer and Couper(2017)]{singer2017somemethodological}
Eleanor Singer and Mick~P Couper.
\newblock Some methodological uses of responses to open questions and other
  verbatim comments in quantitative surveys.
\newblock \emph{Methods, data, analyses: a journal for quantitative methods and
  survey methodology (mda)}, 11\penalty0 (2), 2017.

\bibitem[Blair et~al.(1977)Blair, Sudman, Bradburn, and Stocking]{blair1977ask}
Ed~Blair, Seymour Sudman, Norman~M Bradburn, and Carol Stocking.
\newblock How to ask questions about drinking and sex: Response effects in
  measuring consumer behavior.
\newblock \emph{Journal of marketing Research}, 14\penalty0 (3), 1977.

\bibitem[Lazarsfeld(1935)]{lazarsfeld1935artofaskingwhy}
Paul~F Lazarsfeld.
\newblock The art of asking {WHY} in marketing research: three principles
  underlying the formulation of questionnaires.
\newblock \emph{National marketing review}, 1935.

\bibitem[Dohrenwend(1965)]{dohrenwend1965some}
Barbara~Snell Dohrenwend.
\newblock Some effects of open and closed questions on respondents' answers.
\newblock \emph{Human Organization}, 24\penalty0 (2), 1965.

\bibitem[Schuman and Presser(1979)]{schuman1979open}
Howard Schuman and Stanley Presser.
\newblock The open and closed question.
\newblock \emph{American sociological review}, 1979.

\bibitem[Wenemark et~al.(2010)Wenemark, Hollman~Frisman, Svensson, and
  Kristenson]{wenemark2010respondent}
Marika Wenemark, Gunilla Hollman~Frisman, Tommy Svensson, and Margareta
  Kristenson.
\newblock Respondent satisfaction and respondent burden among differently
  motivated participants in a health-related survey.
\newblock \emph{Field Methods}, 22\penalty0 (4), 2010.

\bibitem[O'Cathain and Thomas(2004)]{o2004any}
Alicia O'Cathain and Kate~J Thomas.
\newblock "any other comments?" {O}pen questions on questionnaires--a bane or a
  bonus to research?
\newblock \emph{BMC medical research methodology}, 4\penalty0 (1), 2004.

\bibitem[Macskassy et~al.(1998)Macskassy, Banerjee, Davison, and
  Hirsh]{macskassy1998human}
Sofus~A Macskassy, Arunava Banerjee, Brian~D Davison, and Haym Hirsh.
\newblock Human {P}erformance on {C}lustering {W}eb {P}ages: {A} {P}reliminary
  {S}tudy.
\newblock In \emph{Proceedings of the 4th {SIGKDD} {C}onference on {K}nowledge
  {D}iscovery and {D}ata {M}ining}, 1998.

\bibitem[Adams(1959)]{adams1959office}
Scott Adams.
\newblock The {O}ffice of {S}cience {I}nformation {S}ervices, {N}ational
  {S}cience {F}oundation.
\newblock \emph{Bulletin of the Medical Library Association}, 47\penalty0 (4),
  1959.

\bibitem[Sager(1967)]{sager1967syntactic}
Naomi Sager.
\newblock Syntactic analysis of natural language.
\newblock In \emph{Advances in computers}, volume~8. Elsevier, 1967.

\bibitem[Grishman and Sundheim(1996)]{grishman1996message}
Ralph Grishman and Beth Sundheim.
\newblock Message understanding conference-6: A brief history.
\newblock In \emph{The 16th International Conference on Computational
  Linguistics (COLING)}, 1996.

\bibitem[Grishman(1997)]{grishman1997information}
Ralph Grishman.
\newblock Information extraction: Techniques and challenges.
\newblock In \emph{International Summer School on Information Extraction}.
  Springer, 1997.

\bibitem[Cowie and Wilks(2000)]{cowie2000information}
Jim Cowie and Yorick Wilks.
\newblock Information extraction.
\newblock \emph{Handbook of Natural Language Processing}, 56, 2000.

\bibitem[Aggarwal and Zhai(2012)]{aggarwal2012mining}
Charu~C Aggarwal and ChengXiang Zhai.
\newblock \emph{Mining text data}.
\newblock Springer Science \& Business Media, 2012.

\bibitem[Blei and Lafferty(2009)]{blei2009topic}
David~M Blei and John~D Lafferty.
\newblock Topic models.
\newblock In \emph{Text Mining}. Chapman and Hall/CRC, 2009.

\bibitem[Roberts et~al.(2014)Roberts, Stewart, Tingley, Lucas, Leder-Luis,
  Gadarian, Albertson, and Rand]{roberts2014structural}
Margaret~E Roberts, Brandon~M Stewart, Dustin Tingley, Christopher Lucas,
  Jetson Leder-Luis, Shana~Kushner Gadarian, Bethany Albertson, and David~G
  Rand.
\newblock Structural topic models for open-ended survey responses.
\newblock \emph{American Journal of Political Science}, 58\penalty0 (4), 2014.

\bibitem[Fitzpatrick(2011)]{fitzpatrick2011humanitiesdonedigitally}
Kathleen Fitzpatrick.
\newblock The {H}umanities, {D}one {D}igitally.
\newblock \emph{The Chronicle of Higher Education}, 2011.

\bibitem[Moretti(2013)]{moretti2013distant}
Franco Moretti.
\newblock \emph{Distant reading}.
\newblock Verso Books, 2013.

\bibitem[Da(2019{\natexlab{a}})]{da2019debacle}
Nan~Z. Da.
\newblock The {D}igital {H}umanities {D}ebacle---{C}omputational methods
  repeatedly come up short.
\newblock \emph{The Chronicle of Higher Education}, 2019{\natexlab{a}}.

\bibitem[Underwood(2019)]{underwood2019dearhumanists}
Ted Underwood.
\newblock Dear {H}umanists: {F}ear {N}ot the {D}igital {R}evolution.
\newblock \emph{The Chronicle of Higher Education}, 2019.

\bibitem[Da(2019{\natexlab{b}})]{da2019computationalcase}
Nan~Z. Da.
\newblock The {C}omputational {C}ase against {C}omputational {L}iterary
  {S}tudies.
\newblock \emph{Critical Inquiry}, 45\penalty0 (3), 2019{\natexlab{b}}.

\bibitem[Karlgren(2019)]{karlgren2019lexical}
Jussi Karlgren.
\newblock How {L}exical {G}old {S}tandards {H}ave {E}ffects {O}n {T}he
  {U}sefulness {O}f {T}ext {A}nalysis {T}ools {F}or {D}igital {S}cholarship.
\newblock In \emph{Proceedings of the Conference and Labs of the Evaluation
  Forum (CLEF)}, 2019.

\bibitem[Georg(2019)]{georg2019virtual}
Carina Georg.
\newblock \emph{Virtual patients in nursing education: teaching, learning and
  assessing clinical reasoning skills}.
\newblock PhD thesis, Karolinska Institutet, Stockholm, 2019.

\bibitem[Sahlgren et~al.(2016)Sahlgren, Gyllensten, Espinoza, Hamfors,
  Karlgren, Olsson, Persson, Viswanathan, and Holst]{sahlgren2016gavagai}
Magnus Sahlgren, Amaru~Cuba Gyllensten, Fredrik Espinoza, Ola Hamfors, Jussi
  Karlgren, Fredrik Olsson, Per Persson, Akshay Viswanathan, and Anders Holst.
\newblock The {G}avagai living lexicon.
\newblock In \emph{Proceedings of the Language Resources and Evaluation
  Conference (LREC)}. ELRA, 2016.

\bibitem[Espinoza et~al.(2018)Espinoza, Hamfors, Karlgren, Olsson, Persson,
  Hamberg, and Sahlgren]{espinoza2018analysis}
Fredrik Espinoza, Ola Hamfors, Jussi Karlgren, Fredrik Olsson, Per Persson,
  Lars Hamberg, and Magnus Sahlgren.
\newblock Analysis of open answers to survey questions through interactive
  clustering and theme extraction.
\newblock In \emph{Proceedings of the Conference on Human Information
  Interaction\&Retrieval (CHIIR)}. ACM, 2018.

\bibitem[Katz(1996)]{katz1996distribution}
Slava~M Katz.
\newblock Distribution of content words and phrases in text and language
  modelling.
\newblock \emph{Natural language engineering}, 2\penalty0 (1), 1996.

\bibitem[Madsen et~al.(2005)Madsen, Kauchak, and Elkan]{madsen2005modeling}
Rasmus~E Madsen, David Kauchak, and Charles Elkan.
\newblock Modeling word burstiness using the dirichlet distribution.
\newblock In \emph{Proceedings of the 22nd {I}nternational {C}onference on
  Machine learning (ICML)}. ACM, 2005.

\bibitem[Luhn(1957)]{luhn1957statistical}
Hans~Peter Luhn.
\newblock A statistical approach to mechanized encoding and searching of
  literary information.
\newblock \emph{IBM Journal of research and development}, 1\penalty0 (4), 1957.

\bibitem[Liu(2012)]{liu2012sentiment}
Bing Liu.
\newblock Sentiment analysis and opinion mining.
\newblock \emph{Synthesis lectures on human language technologies}, 5\penalty0
  (1), 2012.

\bibitem[Darwin(1872)]{darwin1872}
Charles Darwin.
\newblock \emph{The {E}xpression of the {E}motions in {M}an and {A}nimals}.
\newblock John Murray, London, 1872.

\bibitem[James(1884)]{james1884}
W.~James.
\newblock What is an emotion?
\newblock \emph{Mind}, 1884.

\bibitem[Mehrabian and Russell(1974)]{meh1974}
Alfred Mehrabian and James~A Russell.
\newblock \emph{An approach to environmental psychology}.
\newblock M.I.T. Press, 1974.

\bibitem[Morgan and Heise(1988)]{morganheise1988}
Rick~L. Morgan and David Heise.
\newblock {Structure of Emotions}.
\newblock \emph{Social Psychology Quarterly}, 51\penalty0 (1), 1988.

\bibitem[Ekman(1992)]{ekman1992}
Paul Ekman.
\newblock An argument for basic emotions.
\newblock \emph{Cognition and Emotion}, 1992.

\bibitem[Kuppens et~al.(2004)Kuppens, van Mechelen, Smits, and
  de~Boeck]{kuppensEA2004}
Peter Kuppens, Iven van Mechelen, Dirk J.~M. Smits, and Paul de~Boeck.
\newblock Associations {B}etween {E}motions: {C}orrespondence {A}cross
  {D}ifferent {T}ypes of {D}ata and {C}omponential {B}asis.
\newblock \emph{European Journal of Personality}, 18, 2004.

\bibitem[Coan and Allen(2007)]{handbook2007}
J.~A. Coan and J.~J.~B. Allen, editors.
\newblock \emph{{The Handbook of Emotion Elicitation and Assessment}}.
\newblock Oxford University Press, 2007.

\bibitem[Izard and King(2009)]{diff2009}
C.~E. Izard and K.~A. King.
\newblock Differential emotions theory.
\newblock In K.~Scherer, editor, \emph{Oxford Companion to the Affective
  Sciences}. Oxford University Press, 2009.

\bibitem[Rangel et~al.(2013)Rangel, Rosso, Koppel, Stamatatos, and
  Inches]{rangel2013overview}
Francisco Rangel, Paolo Rosso, Moshe Koppel, Efstathios Stamatatos, and Giacomo
  Inches.
\newblock Overview of the author profiling task at {PAN} 2013.
\newblock In \emph{Proceedings of the Conference and Labs of the Evaluation
  Forum (CLEF)}, 2013.

\end{thebibliography}

\end{document}